\documentclass[11pt]{article}

\usepackage[margin=1in]{geometry}
\usepackage[T1]{fontenc}
\usepackage[utf8]{inputenc}
\usepackage{lmodern}
\usepackage{amsmath,amssymb}
\usepackage{graphicx}
\usepackage{booktabs}
\usepackage{microtype}
\usepackage{caption}
\usepackage[hidelinks]{hyperref}

\newcommand{\Lwave}{\mathcal{L}_{\mathrm{wave}}}
\newcommand{\LWhittle}{\mathcal{L}_{\mathrm{Whittle}}}
\newcommand{\Ldyn}{\mathcal{L}_{\mathrm{dyn}}}
\newcommand{\Lsrc}{\mathcal{L}_{\mathrm{src}}}
\newcommand{\Lres}{\mathcal{L}_{\mathrm{res}}}
\newcommand{\Lband}{\mathcal{L}_{\mathrm{band}}}
\newcommand{\pmsd}[2]{#1\,$\pm$\,#2}
\newcommand{\bpmsd}[2]{\textbf{#1\,$\pm$\,#2}}

\title{PhyMind: Physics-Informed Masked Autoencoding for EEG Foundation Models}
\author{A Adith\thanks{Code: \url{https://github.com/Aditharavind/PhyMind}. Weights: \texttt{Aditharavind/PhyMind-*} on the Hugging Face Hub.}\\
\texttt{aditharavind03@gmail.com}}
\date{November 2026}

\begin{document}
\maketitle

\begin{abstract}
EEG foundation models are pre-trained almost exclusively with masked reconstruction through a free-form decoder, sometimes with hand-crafted spectral targets added, as in MANAS-2. We study what happens when that decoder is replaced by a \emph{generative forward model of EEG}. PhyMind keeps a standard masked transformer encoder, but its decoder must explain the masked signal through 68 cortical regions, each driven by damped stochastic oscillators and an aperiodic Ornstein--Uhlenbeck process, projected to the recording's own montage by a fixed boundary-element lead field evaluated per derivation. Training combines masked reconstruction with a Whittle spectral likelihood, an oscillator-dynamics residual, a minimum-norm source prior and a penalised neural residual. In a controlled comparison (same encoder, data, steps and seed), the physics decoder produces encoder tokens whose spectral content is far more linearly decodable than with a plain masked-autoencoder decoder or with the MANAS-2 objective: mean $R^2$ on a token-level spectral probe with amplitude-normalised targets is 0.67/0.77/0.77 on three datasets, against 0.24/0.50/0.50 (MAE) and 0.48/0.67/0.66 (MANAS-2 objective), approaching the released 92.8M-parameter MANAS-2 (0.71/0.81/0.80). Downstream gains in the controlled setting are small (four-task mean balanced accuracy 56.6 vs.\ 53.8 for MAE frozen, 63.8 vs.\ 62.7 fine-tuned). Region-level features from the physics decoder raise frozen four-class motor-imagery accuracy by 7--14 points, so a 7.2M-parameter model pre-trained on 3k hours of open EEG is comparable to the released MANAS-2 with frozen features (63.0 vs.\ 61.6). After full fine-tuning, MANAS-2 remains better (66.8 vs.\ 65.4 for our best model). The inferred sources do not reproduce the lateralisation that a classical inverse solution finds in motor imagery, so we do not claim they are anatomically interpretable. Code and weights are released.
\end{abstract}

\section{Introduction}
Self-supervised pre-training has produced a family of EEG foundation models (BENDR~\cite{kostas2021bendr}, BIOT~\cite{yang2023biot}, LaBraM~\cite{jiang2024labram}, CBraMod~\cite{wang2025cbramod}, REVE~\cite{ouahidi2025reve}, MANAS-2~\cite{kulkarni2026manas2}). Most follow the masked-autoencoder recipe of vision~\cite{he2022mae}: hide some patches, encode the rest, and decode the hidden waveform with an unconstrained network. MANAS-2 adds six log band-power targets per patch (Raw-Band Hybrid, RBH) and a physics-motivated RMS continuity penalty at patch boundaries (ConRec).

The generative process of scalp EEG is unusually well understood. Cortical current sources oscillate in characteristic bands on top of an aperiodic background~\cite{donoghue2020specparam}, and they reach the electrodes through volume conduction, which is linear, instantaneous and computable from a head model~\cite{gramfort2013mne,hamalainen1994mne}. Reconstruction-trained embeddings have been reported to be dominated by the aperiodic component and by subject identity~\cite{kommineni2026aperiodic,lin2026identity}, and in most models electrode geometry only enters as a positional embedding.

\begin{figure}[t]
  \centering
  \includegraphics[width=\textwidth]{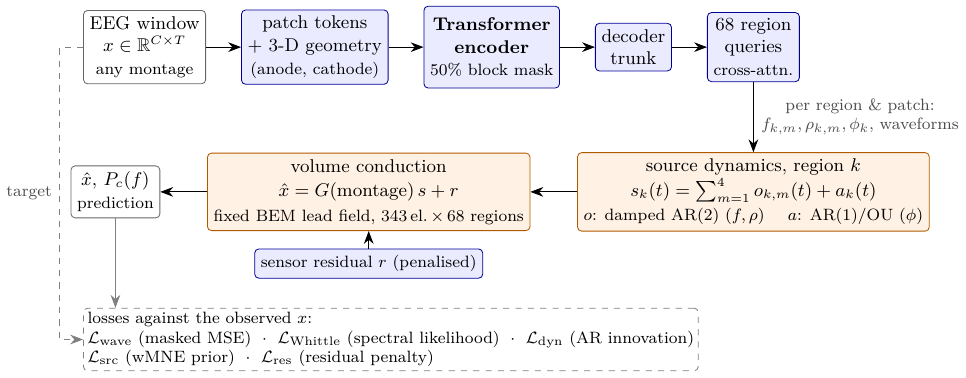}
  \caption{PhyMind. Blue: learned modules; orange: fixed physics. The encoder is a standard masked transformer. The decoder infers, for each of 68 cortical regions and each 1\,s patch, oscillator and aperiodic parameters and waveforms, which a fixed BEM lead field, evaluated for the recording's own derivations, projects to the sensors. Only a small, penalised sensor residual $r$ bypasses the physics.}
  \label{fig:overview}
\end{figure}

\begin{figure}[t]
  \centering
  \includegraphics[width=\textwidth]{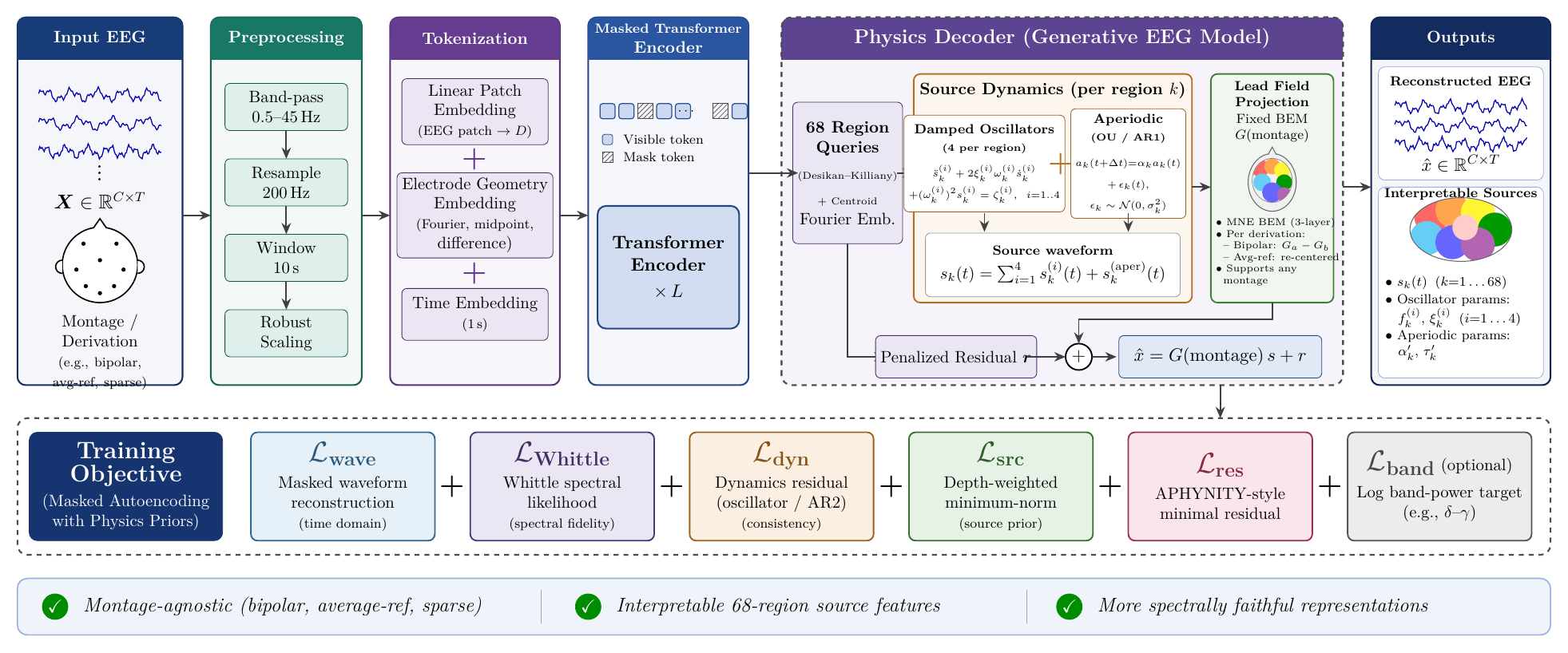}
  \caption{Detailed PhyMind architecture: preprocessing and tokenisation, masked transformer encoder, physics decoder (68 region queries, oscillator and aperiodic source dynamics, fixed BEM lead-field projection per derivation, penalised residual), outputs, and the training objective.}
  \label{fig:arch}
\end{figure}

We ask a narrow question: what does replacing the free-form decoder with a physical forward model change in the encoder? PhyMind's decoder is a parametric, differentiable generative model of EEG (Fig.~\ref{fig:overview}; a detailed architecture diagram is given in Fig.~\ref{fig:arch}). The encoder must produce tokens from which a small network can infer cortical source dynamics that, after a fixed lead field, explain the masked channels and their spectra. We answer the question with a controlled comparison against a plain MAE decoder and the MANAS-2 objective, and we report a comparison with the released MANAS-2 for context, including where it is better.

\paragraph{Contributions.}
\begin{itemize}
  \item A physics-informed masked-autoencoding objective for EEG: a fixed BEM lead field evaluated per derivation (any montage or reference), oscillator and aperiodic source dynamics, a Whittle spectral likelihood, and a constrained neural residual.
  \item A controlled comparison of MAE, the MANAS-2 objective (RBH + ConRec) and PhyMind with identical encoder, data, sampler, steps and seed, under frozen probing, full fine-tuning and a token-level spectral probe whose targets are normalised so that models with per-channel-normalised input are not penalised.
  \item The main empirical finding: the physics decoder makes spectral structure markedly more linearly accessible in encoder tokens (mean $R^2$ +0.27 to +0.43 over MAE, +0.10 to +0.19 over the MANAS-2 objective, without any band-power target), while downstream accuracy gains are small.
  \item Region-level source features that improve frozen motor-imagery decoding, together with a negative result: the sources do not localise motor imagery the way a classical minimum-norm inverse with the same lead field does.
  \item An exact, padding-aware re-implementation of the released MANAS-2 encoder that loads its weights strictly (bit-identical outputs), used for evaluation and for a retrofit experiment (PhyMind-M2), which is negative for classification.
\end{itemize}

\section{Related work}
\paragraph{EEG foundation models.} BENDR~\cite{kostas2021bendr} used contrastive learning; BIOT~\cite{yang2023biot} tokenises channels independently; LaBraM~\cite{jiang2024labram} predicts neural codes from a learned spectral tokenizer; CBraMod~\cite{wang2025cbramod} uses criss-cross spatial/temporal attention; REVE~\cite{ouahidi2025reve} introduced a 4-D Fourier positional encoding of electrode position and time and pre-trains a masked autoencoder on more than 60k hours. MANAS-2~\cite{kulkarni2026manas2} adopts this positional encoding and adds the RBH and ConRec targets. DeeperBrain~\cite{wang2026deeperbrain} uses distance-based channel embeddings and neuro-statistical targets. BrainOmni~\cite{xiao2025brainomni} learns latent ``sources'' shared by EEG and MEG, but they are not tied to a lead field. To our knowledge none of these decodes through a biophysical forward model.

\paragraph{Physics and EEG.} Neural-mass models~\cite{jansen1995} combined with lead fields underpin supervised source imaging (DeepSIF~\cite{sun2022deepsif}). NeuroDyn-EEG~\cite{cui2026neurodyn} couples a Jansen--Rit model with a lead field, but trains on simulations. PINR-ESI~\cite{zhou2026pinresi} fits a lead-field-constrained implicit representation per recording. Neither is a pre-trained encoder learned from large real corpora.

\paragraph{Hybrid physics--ML.} PINNs~\cite{raissi2019pinn} penalise physical residuals. APHYNITY~\cite{yin2021aphynity} and physics-integrated VAEs~\cite{takeishi2021physvae} add a neural residual to an incomplete physical model and keep it minimal so that the physics explains as much as possible. PhyMind's decoder follows this design.

\section{Method}
\subsection{Tokenisation and encoder}
Recordings are band-passed (0.5--45\,Hz), resampled to 200\,Hz, average-referenced when unipolar, and cut into 10\,s windows. Each window is divided by a single robust scale (median absolute deviation) \emph{shared across channels}. Per-channel $z$-scoring, used by MANAS-2 and most other models, would destroy the relative channel amplitudes the lead field predicts. A channel is represented by its derivation $(a, b)$: an anode and either a cathode or ``average reference'', both indexed into the 343-electrode 10-05 system. Each 1\,s patch is linearly embedded. A Fourier embedding of the electrode coordinates (the anode--cathode midpoint and difference for bipolar channels) and a learned time embedding are added, and a pre-norm transformer encodes all channel$\times$patch tokens jointly with a key-padding mask, so channel counts can vary within a batch.

\paragraph{Masking.} We mask 50\% of valid tokens. Selection is block-structured: random tokens, whole time patches (25\% probability) and whole channels (20\%). Masking whole channels forces the model to predict a channel from its neighbours, which is exactly the problem that volume conduction, and hence the lead field, solves.

\subsection{Physics decoder}
\paragraph{Source space.} We use the 68 Desikan--Killiany regions~\cite{desikan2006} on the FreeSurfer \texttt{fsaverage} template~\cite{fischl2012freesurfer}. A two-layer transformer trunk processes the full token grid (visible tokens plus mask tokens). Learned region queries, augmented with a Fourier embedding of each region's centroid, then cross-attend to the trunk output and produce one token per (region, patch).

\paragraph{Source dynamics.} Each region's activity is a sum of $M{=}4$ damped stochastic oscillators and one aperiodic component,
\begin{equation}
  s_k(t) = \sum_{m=1}^{M} o_{k,m}(t) + a_k(t).
\end{equation}
Each oscillator is the AR(2) discretisation of $\ddot y + 2\gamma\dot y + \omega^2 y = \varepsilon$ (cf.\ Matsuda and Komaki~\cite{matsuda2017}):
\begin{equation}
  o_t = 2\rho\cos(2\pi f/f_s)\,o_{t-1} - \rho^2 o_{t-2} + \varepsilon_t,
\end{equation}
with $f \in [1, 45]$\,Hz and $\rho \in (0, 1)$ predicted per (region, patch). The aperiodic component is an AR(1)/Ornstein--Uhlenbeck process with coefficient $\phi$. Its Lorentzian spectrum is the ``knee'' aperiodic model of specparam~\cite{donoghue2020specparam}, and its slope is related to excitation/inhibition balance~\cite{gao2017ei}. The decoder emits the waveforms $o_{k,m}$ and $a_k$ directly; the dynamics enter as soft constraints (below), not as a hard recursion.

\paragraph{Volume conduction.} Sensor activity is
\begin{equation}
  \hat{x} = G(\text{montage})\, s + r.
\end{equation}
$G \in \mathbb{R}^{343\times 68}$ is a \emph{fixed} lead field, computed once with an MNE~\cite{gramfort2013mne} three-layer BEM on \texttt{fsaverage}. It uses ico-5 cortical dipoles with fixed normal orientation, averaged within each parcel and normalised to unit median column norm. Per sample, a bipolar channel $a{-}b$ uses the row $G_a - G_b$, and average-referenced channels use rows re-centred over the channels actually present. CHB-MIT's double banana, Sleep-EDF's Fpz--Cz/Pz--Oz and 64-channel caps are therefore all explained by one source model. $r$ is a sensor-level residual for non-cortical activity (ocular, muscle, template mismatch); it is zero-initialised and penalised.

\subsection{Objective}
\begin{equation}
  \mathcal{L} = \Lwave + \lambda_W \LWhittle + \lambda_D \Ldyn + \lambda_S \Lsrc + \lambda_R \Lres \;(+\lambda_B \Lband).
\end{equation}
\begin{itemize}
  \item $\Lwave$: mean squared error of $\hat{x}$ on masked tokens.
  \item $\Ldyn$: normalised innovation energy $\|e\|^2/\|o\|^2$ of every component under its AR model, computed over the full 10\,s window. Because the recursion crosses patch boundaries, it acts as a generative continuity constraint and replaces ConRec's boundary heuristic.
  \item $\LWhittle$: the Whittle likelihood~\cite{whittle1953}, in deviance form, of each channel's Hann-windowed periodogram $I_c(f)$ under the model spectrum
  \begin{equation}
    P_c(f) = \sum_k G_{ck}^2 \Big[\sum_m S^{\mathrm{AR2}}_{k,m}(f) + S^{\mathrm{AR1}}_k(f)\Big] + |R_c(f)|^2 + \epsilon_c,
  \end{equation}
  where the spectrum is evaluated at 1--45\,Hz in 1\,Hz bins for each 1\,s patch, the innovation variances are taken from the empirical residuals, and $\epsilon_c = 10^{-2}\,\overline{I_c}$ is a scale-aware floor (1\% of the channel's mean periodogram, gradient-detached) that keeps the term bounded while $\hat{x}$ is still near zero. Periodic peaks and the aperiodic knee have separate parameters.
  \item $\Lsrc = \sum_k w_k \|s_k\|^2$ with $w_k = \|G_k\|^2$: the depth-weighted minimum-norm prior~\cite{hamalainen1994mne}.
  \item $\Lres = \|r\|^2$: keeps the non-physical path minimal~\cite{yin2021aphynity}.
\end{itemize}
Weights: $\lambda_D{=}0.1$, $\lambda_S{=}10^{-3}$, $\lambda_R{=}0.1$, and $\lambda_W{=}0.2$ (v1) or $0.5$ (v2). v2 evaluates the Whittle term on all tokens rather than only masked ones. v2-H additionally adds a linear head on encoder tokens that predicts the six log band powers of each token ($\log(1{+}\cdot)$ of the mean periodogram in each band, standardised per window and band over channels and patches; L1 loss, $\lambda_B{=}0.5$). PhyMind-lite and PhyMind-large use the v2-H objective. Because this head is a direct, supervised band-power target on encoder tokens, we treat models trained with it separately whenever spectral content is probed (Sec.~\ref{sec:spectral}).

\paragraph{Inference.} \texttt{PhyMind.explain} returns, for any montage, 68 regional source time series together with per-second oscillator frequencies, dampings and aperiodic coefficients. These are regularised model quantities, not validated source localisations (Sec.~\ref{sec:localise}).

\subsection{PhyMind-M2: physics decoder on the MANAS-2 encoder}
We re-implemented the MANAS-2 encoder (22 layers, $d{=}512$, 8 heads, 200-sample patches with stride 180, 4-D Fourier positional encoding, and the GEGLU variant used in the release) with a key-padding mask. It loads the released weights strictly; on unpadded input its outputs match the release exactly (max.\ abs.\ difference 0.0), and padding changes outputs by less than $10^{-5}$. MANAS-2 $z$-scores each channel, so we feed it $z$-scored visible samples and add a zero-initialised projection of each channel's log standard deviation. This is intended to restore the amplitude information that the lead field needs. The 11 stride-180 tokens are interpolated to the decoder's 10 one-second patches. PhyMind-M2 continues pre-training of this encoder (learning rate $\times 0.2$) with the physics decoder. M2+RBH is the identical control trained with the MANAS-2 objective.

\section{Experimental setup}
\label{sec:setup}
\paragraph{Pre-training data.} All data are openly available. Phase~1 uses 621\,h: CHB-MIT~\cite{goldberger2000physionet,shoeb2009} (bipolar, 23 channels), a subset of Sleep-EDF cassette~\cite{kemp2000sleepedf} (2 bipolar channels), and four OpenNeuro~\cite{markiewicz2021openneuro} resting-state and oddball datasets (ds002778, ds003478, ds003490, ds004584; 32--64 channels). The final runs (lite, large, M2) use 2,959\,h from the same sources with all of Sleep-EDF cassette. The three dataset groups (CHB-MIT, Sleep-EDF, OpenNeuro) are sampled with equal total mass, so the small OpenNeuro group is heavily over-sampled relative to its size. At most 32 channels are drawn per window, and channels are randomly dropped (probability 0.2).

\paragraph{Models.} Pre-norm transformers with drop-path 0.1. lite: $d{=}256$, 6 layers (4.9M encoder parameters, 7.2M with decoder); base: $d{=}384$, 12 layers; large: $d{=}512$, 16 layers (50.8M / 59.6M). AdamW (lr $3\times10^{-4}$, weight decay 0.05, cosine schedule), bf16, batch 128, one GPU. Phase~1 base models (MAE, RBH, v1, v2, v2-H) train for 15k steps with seed 0; lite and large train for 20k and 30k steps. The MAE baseline uses the same encoder with a transformer decoder that regresses masked patches. RBH adds MANAS-2's six log band-power targets (weight 0.5) and ConRec ($q{=}32$, $\lambda{=}3$). There is one pre-training seed per model.

\paragraph{Downstream tasks.} EEGMAT~\cite{zyma2019eegmat} (mental arithmetic vs.\ rest, 36 subjects, 19 ch.); PhysioNet motor imagery~\cite{schalk2004bci2000} (4 classes: left fist, right fist, both fists, both feet; 105 subjects, 64 ch., 4\,s); ADFTD~\cite{miltiadous2023adftd} (Alzheimer's disease / frontotemporal dementia / control, 88 subjects, 19 ch.); and Sleep-EDF sleep-telemetry~\cite{kemp2000sleepedf} (5 stages, 22 subjects, 2 ch., 30\,s epochs; both nights of a subject share one subject ID). No downstream subject is in the pre-training data: ADFTD is ds004504, which is not one of our OpenNeuro pre-training sets, and sleep-telemetry is disjoint from the cassette recordings. PhyMind's learned time embedding was pre-trained on 10 one-second patches; the 30-patch Sleep-ST epochs therefore use positions never seen in pre-training, which, if anything, disadvantages PhyMind on that task.

\paragraph{Protocols.} All protocols use the same three subject-disjoint 60/20/20 splits (seeds 0--2); we report mean $\pm$ sd balanced accuracy. Because the training sets of the splits overlap, the sd understates uncertainty. \emph{LP-Avg}: frozen encoder, mean-pooled tokens, multinomial logistic regression with $C$ selected on validation balanced accuracy. For PhyMind we additionally report one pre-declared variant, ``+ src'', which concatenates the pooled tokens with the decoder's 68 per-region log powers and mean oscillator and aperiodic parameters. \emph{Fine-tuning}: encoder plus a LayerNorm--linear head trained end to end with AdamW (lr $5\times10^{-5}$ for the encoder, $\times10$ for the head, weight decay 0.05, cosine schedule with one warm-up epoch), class-weighted cross-entropy with label smoothing 0.1, batch 32, bf16, for 30/15/15/4 epochs (EEGMAT/MI/ADFTD/Sleep-ST); the epoch with the best validation balanced accuracy is evaluated on test. MANAS-2 is fine-tuned with the same recipe from its released weights, with per-channel $z$-scored input as in its model card. No hyper-parameters were tuned per model.

\paragraph{Spectral probe.} A ridge regression is fit on frozen tokens (2\% random token subsample) to predict, for each token's channel and 1\,s patch: six log band powers ($\log(1{+}\cdot)$ of the mean Hann periodogram in 0.5--4, 4--8, 8--13, 13--20, 20--30, 30--45\,Hz), alpha flow (the patch-to-patch change in alpha log power; the first patch is excluded), the aperiodic exponent (least-squares log--log slope over 2--40\,Hz) and periodic alpha power (largest residual above that fit in 8--13\,Hz). Targets are computed from the per-channel $z$-scored window, so absolute channel amplitude, which MANAS-2's normalised input cannot contain, does not enter any target. The ridge penalty is selected on validation subjects and $R^2$ reported on test subjects. Targets follow each encoder's own patch grid (stride 180 for MANAS-2). This is our probe, inspired by but not identical to MANAS-2's analysis.

\paragraph{Model selection.} The PhyMind variants (v1$\to$v2$\to$v2-H$\to$lite/large) were developed sequentially while inspecting LP and probe results on these same tasks and splits; no separate model-selection benchmark was held out. The controlled block (MAE, RBH, v1) was fixed before this iteration.

\section{Results}
\subsection{Spectral content of encoder tokens}
\label{sec:spectral}
\begin{figure}[t]
  \centering
  \includegraphics[width=\textwidth]{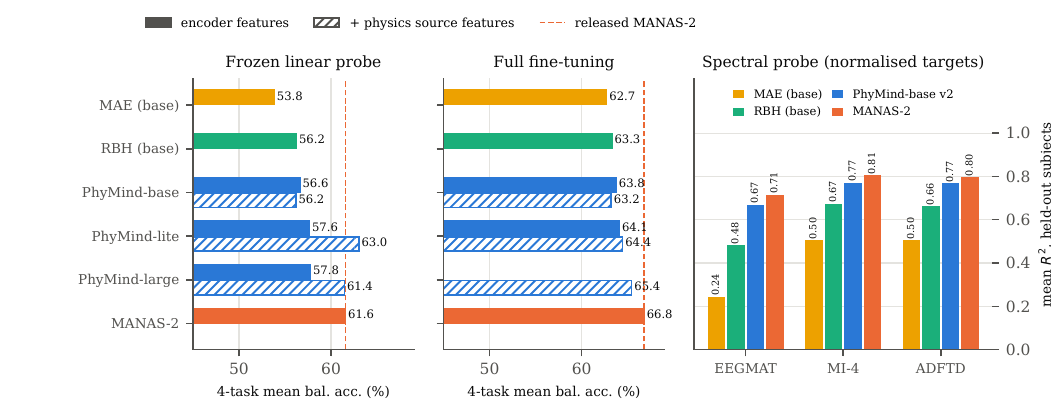}
  \caption{Left, middle: four-task mean balanced accuracy with frozen features and after full fine-tuning (hatched: + physics source features). Dashed line: released MANAS-2. Right: spectral-probe mean $R^2$ with amplitude-normalised targets for models without any band-power target on encoder tokens.}
  \label{fig:results}
\end{figure}

Table~\ref{tab:probe} and Fig.~\ref{fig:results} (right) show the clearest effect of the physics decoder. With encoder, data, steps and seed fixed, PhyMind-base v2 reaches mean $R^2$ 0.67/0.77/0.77 on EEGMAT/MI/ADFTD, against 0.24/0.50/0.50 for MAE and 0.48/0.67/0.66 for the MANAS-2 objective (RBH), although RBH receives band power as an explicit target and v2 does not. The gain is broad: on EEGMAT, $\beta$ and $\gamma$ power, the aperiodic exponent and periodic alpha all improve by 0.2--0.8 over MAE. The strength of the spectral likelihood matters: v1, which applies the Whittle term only on masked tokens with weight 0.2, lies between MAE and RBH (0.40/0.59/0.60). The released MANAS-2 remains slightly ahead of v2 (0.71/0.81/0.80) with a much larger model and corpus. Lite and large exceed it (0.80--0.88), but they were trained with a band-power head on encoder tokens, so their probe numbers are largely a consequence of direct supervision and are reported for completeness only.

\begin{table}[t]
\centering
\caption{Frozen linear probing (LP-Avg): balanced accuracy (\%), mean $\pm$ sd over three subject-disjoint splits. ``+ src'' (pre-declared feature set) concatenates pooled encoder tokens with the physics decoder's 68-region source features (PhyMind only). Params include the decoder for PhyMind (needed for + src); encoder-only counts are 4.9M (lite), 50.8M (large) and 92.8M (MANAS-2). $^\dagger$~continued pre-training of the released MANAS-2 encoder. $^*$~not stated in the MANAS-2 model card.}
\label{tab:lp}
\small
\setlength{\tabcolsep}{4pt}
\begin{tabular}{lrrccccc}
\toprule
Model & Params & Hours & EEGMAT & MI-4 & ADFTD & Sleep-ST & Mean \\
\midrule
MAE                     & 23.4M & 621 & \pmsd{64.3}{5.6} & \pmsd{31.0}{0.5} & \pmsd{50.5}{6.6} & \pmsd{69.4}{1.2} & 53.8 \\
RBH (MANAS-2 obj.)      & 23.4M & 621 & \pmsd{69.9}{1.9} & \pmsd{29.9}{1.0} & \pmsd{52.7}{9.2} & \pmsd{72.5}{1.7} & 56.2 \\
PhyMind-base v1         & 25.6M & 621 & \pmsd{71.4}{0.6} & \pmsd{29.1}{1.0} & \pmsd{54.2}{9.1} & \pmsd{71.8}{1.8} & 56.6 \\
PhyMind-base v1 + src   & 25.6M & 621 & \pmsd{70.5}{3.0} & \pmsd{29.7}{0.6} & \pmsd{52.5}{9.3} & \pmsd{72.0}{1.8} & 56.2 \\
PhyMind-base v2         & 25.6M & 621 & \pmsd{67.3}{4.8} & \pmsd{28.7}{1.5} & \pmsd{51.7}{8.7} & \pmsd{73.7}{0.6} & 55.4 \\
PhyMind-base v2 + src   & 25.6M & 621 & \pmsd{65.3}{0.6} & \pmsd{32.4}{1.1} & \pmsd{51.7}{10.5} & \pmsd{73.9}{0.7} & 55.8 \\
PhyMind-base v2-H       & 25.6M & 621 & \pmsd{65.4}{2.6} & \pmsd{30.6}{0.7} & \pmsd{55.9}{8.6} & \pmsd{74.3}{1.0} & 56.6 \\
PhyMind-base v2-H + src & 25.6M & 621 & \pmsd{69.2}{3.5} & \pmsd{34.3}{0.8} & \pmsd{59.6}{2.6} & \pmsd{74.2}{1.0} & 59.3 \\
\midrule
PhyMind-lite            & 7.2M  & 2959 & \pmsd{67.7}{6.6} & \pmsd{33.2}{0.1} & \pmsd{55.9}{8.9} & \pmsd{73.8}{1.0} & 57.6 \\
PhyMind-lite + src      & 7.2M  & 2959 & \pmsd{73.4}{3.4} & \pmsd{47.0}{1.0} & \pmsd{57.3}{6.8} & \pmsd{74.4}{0.8} & 63.0 \\
PhyMind-large           & 59.6M & 2959 & \pmsd{65.2}{6.7} & \pmsd{36.1}{0.3} & \pmsd{54.4}{9.1} & \pmsd{75.3}{1.0} & 57.8 \\
PhyMind-large + src     & 59.6M & 2959 & \pmsd{69.4}{2.6} & \pmsd{45.2}{0.1} & \pmsd{55.9}{7.7} & \pmsd{75.2}{1.0} & 61.4 \\
\midrule
M2 + RBH (control)      & 95.2M & 2959$^\dagger$ & \pmsd{70.7}{3.0} & \pmsd{43.6}{0.7} & \pmsd{59.4}{9.1} & \pmsd{75.3}{1.4} & 62.2 \\
PhyMind-M2              & 97.4M & 2959$^\dagger$ & \pmsd{71.2}{5.0} & \pmsd{37.9}{0.8} & \pmsd{55.9}{8.5} & \pmsd{74.9}{1.3} & 60.0 \\
PhyMind-M2 + src        & 97.4M & 2959$^\dagger$ & \pmsd{71.1}{5.8} & \pmsd{43.6}{0.3} & \pmsd{51.9}{6.2} & \pmsd{75.2}{1.3} & 60.5 \\
MANAS-2 (released)      & 92.8M & n/s$^*$ & \pmsd{72.2}{3.4} & \pmsd{40.5}{0.6} & \pmsd{57.9}{6.1} & \pmsd{75.6}{1.5} & 61.6 \\
\bottomrule
\end{tabular}
\end{table}

\subsection{Controlled downstream comparison}
In the controlled block, frozen four-task means are 53.8 (MAE), 56.2 (RBH) and 56.6 (PhyMind v1) (Table~\ref{tab:lp}); after full fine-tuning they are 62.7, 63.3 and 63.8 (Table~\ref{tab:ft}). The physics objective is therefore never worse than either baseline on average, but the margins are small relative to the split-to-split sd (up to 9--10 points on EEGMAT and ADFTD), and per-task rankings vary. v2, the variant with the strongest spectral representation, is not better than v1 for classification (55.4 frozen). We conclude that more faithful spectral tokens do not by themselves translate into higher downstream accuracy on these tasks.

\subsection{Source features}
\label{sec:srcfeat}
The pre-declared ``+ src'' features help most on four-class motor imagery with frozen encoders: from 33.2 to 47.0 (lite) and from 36.1 to 45.2 (large), against 40.5 for the released MANAS-2; these gains exceed the split sd ($\le$1 point) by a wide margin. On the other tasks the effect is within noise, and for the phase-1 base models it is small or absent. After full fine-tuning the source features add at most 1.6 points (lite, MI), because the encoder itself can adapt.

\begin{table}[t]
\centering
\caption{Full fine-tuning (encoder and linear head trained end to end): balanced accuracy (\%), mean $\pm$ sd over the same three subject-disjoint splits as Table~\ref{tab:lp}; the epoch with the best validation balanced accuracy is evaluated. The first block is the controlled comparison (same encoder, data, steps and seed). Bold: best per task.}
\label{tab:ft}
\small
\setlength{\tabcolsep}{4pt}
\begin{tabular}{lrccccc}
\toprule
Model & Params & EEGMAT & MI-4 & ADFTD & Sleep-ST & Mean \\
\midrule
MAE (base)            & 23.4M & \pmsd{65.6}{9.3} & \pmsd{55.2}{1.4} & \pmsd{57.4}{2.9} & \pmsd{72.7}{1.0} & 62.7 \\
RBH (base)            & 23.4M & \pmsd{62.8}{6.1} & \pmsd{55.3}{1.5} & \bpmsd{60.5}{3.0} & \pmsd{74.5}{1.8} & 63.3 \\
PhyMind-base v1       & 25.6M & \pmsd{69.4}{2.3} & \pmsd{55.4}{1.3} & \pmsd{56.9}{9.8} & \pmsd{73.3}{1.9} & 63.8 \\
PhyMind-base v1 + src & 25.6M & \pmsd{68.5}{4.8} & \pmsd{54.9}{0.8} & \pmsd{56.6}{9.4} & \pmsd{72.9}{1.7} & 63.2 \\
\midrule
PhyMind-lite          & 7.2M  & \pmsd{65.6}{2.3} & \pmsd{57.3}{0.5} & \pmsd{58.3}{9.3} & \pmsd{75.2}{1.5} & 64.1 \\
PhyMind-lite + src    & 7.2M  & \pmsd{67.1}{4.2} & \pmsd{58.9}{0.8} & \pmsd{57.0}{6.5} & \pmsd{74.8}{1.5} & 64.4 \\
PhyMind-large + src   & 59.6M & \pmsd{66.4}{0.5} & \pmsd{60.8}{0.6} & \pmsd{59.5}{2.4} & \pmsd{75.0}{3.1} & 65.4 \\
MANAS-2 (released)    & 92.8M & \bpmsd{73.7}{4.9} & \bpmsd{63.7}{0.9} & \pmsd{52.2}{8.9} & \bpmsd{77.4}{1.8} & 66.8 \\
\bottomrule
\end{tabular}
\end{table}

\subsection{Comparison with the released MANAS-2}
With frozen features, PhyMind-lite + src (63.0) is comparable to MANAS-2 (61.6): it is clearly better on MI (47.0 vs.\ 40.5), and EEGMAT, ADFTD and Sleep-ST are within split variance. With encoder features alone, MANAS-2 is better (61.6 vs.\ 57.6--57.8). After full fine-tuning, MANAS-2 is better on average (66.8 vs.\ 65.4 for PhyMind-large + src and 64.4 for lite + src) and on three of four tasks (EEGMAT 73.7 vs.\ 67.1, MI 63.7 vs.\ 60.8, Sleep-ST 77.4 vs.\ 75.2); PhyMind is ahead on ADFTD (59.5 vs.\ 52.2), where MANAS-2's sd is 8.9 points. MANAS-2 has a larger encoder (92.8M vs.\ 4.9--50.8M) and was pre-trained on a substantially larger corpus. Our models are therefore competitive at small scale but do not surpass it.

\subsection{Physics decoder on MANAS-2 (PhyMind-M2)}
Continuing pre-training of the released MANAS-2 encoder on our 2,959\,h for 20k steps is negative for classification. With the RBH objective (control), the frozen four-task mean rises from 61.6 to 62.2; with the physics decoder it falls to 60.0 (60.5 + src), mainly on MI (37.9). We hypothesise that this is a normalisation mismatch: MANAS-2's encoder consumes per-channel $z$-scored input, which removes the relative channel amplitudes the lead field explains, and our log-scale side channel restores them only partially. We have not tested this hypothesis directly.

\section{Analysis}
\subsection{Do the sources localise?}
\label{sec:localise}
\begin{figure}[t]
  \centering
  \includegraphics[width=\textwidth]{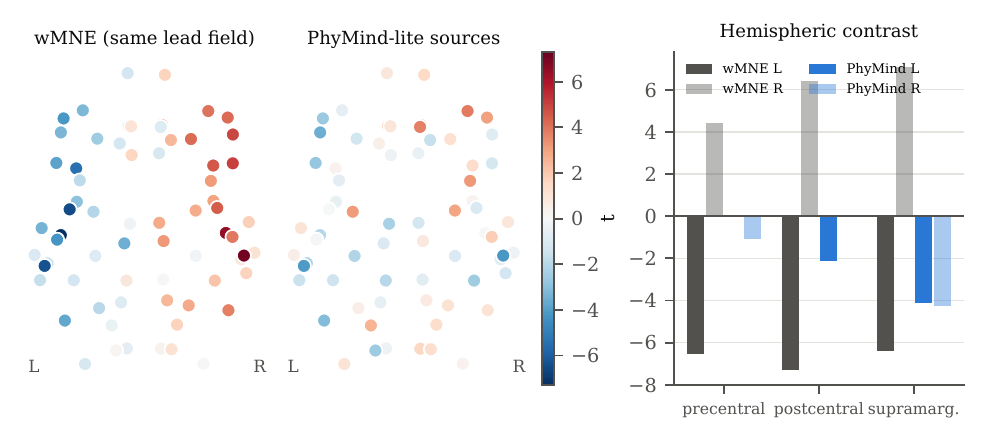}
  \caption{Right- minus left-fist motor imagery, PhysioNet MI (paired $t$ over subjects). Left: classical depth-weighted minimum-norm estimate (wMNE) with the same lead field. Middle: PhyMind-lite source power. Right: hemispheric contrast for sensorimotor regions. wMNE shows the expected contralateral pattern; PhyMind's sources do not.}
  \label{fig:localisation}
\end{figure}

As a sanity check we contrasted right- and left-fist imagery on PhysioNet MI (Fig.~\ref{fig:localisation}). At the sensors the expected contralateral alpha/beta desynchronisation is present (C3 $t{=}{-}5.5$, C4 $t{=}{+}6.5$), and a classical wMNE inverse with the same lead field recovers it in source space (precentral $t{=}{-}6.5$ left / $+4.4$ right; postcentral $-7.3$ / $+6.4$). PhyMind-lite's inferred sources do not (precentral $-0.03$ / $-1.1$; postcentral $-2.1$ / $-0.05$). The decoder's 68 ``regions'' therefore behave as a learned, lead-field-constrained basis that carries task information (Sec.~\ref{sec:srcfeat}), not as anatomically faithful source estimates. We do not claim anatomical interpretability.

\begin{table}[t]
\centering
\caption{Frozen token-level spectral probe (ridge regression, held-out subjects, $R^2$, mean of three splits). Targets are computed from the per-channel $z$-scored window, so absolute channel amplitude, which MANAS-2's normalised input cannot contain, is removed from every target (definitions in Sec.~\ref{sec:setup}). $^\ddagger$~Pre-trained with an auxiliary linear head predicting log band power from encoder tokens, i.e.\ directly supervised on six of the nine targets; these rows are not a fair comparison and are excluded from bolding. Bold: best among the remaining models.}
\label{tab:probe}
\small
\setlength{\tabcolsep}{4.5pt}
\begin{tabular}{lcccccccccc}
\toprule
Model & $\delta$ & $\theta$ & $\alpha$ & $\beta_1$ & $\beta_2$ & $\gamma$ & $\alpha$-flow & aper. & $\alpha$-per. & mean \\
\midrule
\multicolumn{11}{l}{\emph{EEGMAT}} \\
MAE                 & 0.53 & 0.37 & 0.65 & 0.12 & 0.01 & -0.32 & 0.26 & 0.03 & 0.55 & 0.24 \\
RBH (MANAS-2 obj.)  & 0.60 & 0.48 & 0.85 & 0.43 & 0.40 & 0.01 & \textbf{0.41} & 0.38 & 0.79 & 0.48 \\
PhyMind-base v1     & 0.54 & 0.34 & 0.74 & 0.22 & 0.34 & 0.13 & 0.29 & 0.34 & 0.69 & 0.40 \\
PhyMind-base v2     & 0.66 & 0.64 & \textbf{0.87} & 0.75 & 0.71 & 0.52 & 0.39 & 0.66 & \textbf{0.83} & 0.67 \\
PhyMind-lite$^\ddagger$  & 0.94 & 0.91 & 0.95 & 0.88 & 0.83 & 0.65 & 0.41 & 0.79 & 0.87 & 0.80 \\
PhyMind-large$^\ddagger$ & 0.95 & 0.93 & 0.95 & 0.90 & 0.85 & 0.71 & 0.43 & 0.88 & 0.89 & 0.83 \\
MANAS-2 (released)  & \textbf{0.77} & \textbf{0.80} & 0.87 & \textbf{0.76} & \textbf{0.75} & \textbf{0.66} & 0.33 & \textbf{0.71} & 0.79 & \textbf{0.71} \\
\midrule
\multicolumn{11}{l}{\emph{MI-4}} \\
MAE                 & 0.69 & 0.58 & 0.62 & 0.34 & 0.47 & 0.49 & 0.32 & 0.59 & 0.45 & 0.50 \\
RBH (MANAS-2 obj.)  & 0.75 & 0.65 & 0.82 & 0.55 & 0.67 & 0.66 & \textbf{0.46} & 0.79 & 0.73 & 0.67 \\
PhyMind-base v1     & 0.68 & 0.53 & 0.70 & 0.44 & 0.61 & 0.70 & 0.32 & 0.76 & 0.59 & 0.59 \\
PhyMind-base v2     & 0.77 & 0.73 & 0.87 & 0.80 & 0.82 & 0.82 & 0.44 & 0.89 & \textbf{0.78} & 0.77 \\
PhyMind-lite$^\ddagger$  & 0.93 & 0.90 & 0.91 & 0.87 & 0.86 & 0.87 & 0.51 & 0.93 & 0.83 & 0.85 \\
PhyMind-large$^\ddagger$ & 0.95 & 0.92 & 0.94 & 0.90 & 0.89 & 0.89 & 0.54 & 0.96 & 0.87 & 0.88 \\
MANAS-2 (released)  & \textbf{0.86} & \textbf{0.85} & \textbf{0.89} & \textbf{0.84} & \textbf{0.86} & \textbf{0.88} & 0.41 & \textbf{0.90} & 0.76 & \textbf{0.81} \\
\midrule
\multicolumn{11}{l}{\emph{ADFTD}} \\
MAE                 & 0.61 & 0.56 & 0.75 & 0.32 & 0.40 & 0.40 & 0.34 & 0.52 & 0.64 & 0.50 \\
RBH (MANAS-2 obj.)  & 0.68 & 0.62 & 0.87 & 0.54 & 0.62 & 0.64 & \textbf{0.44} & 0.73 & 0.82 & 0.66 \\
PhyMind-base v1     & 0.62 & 0.54 & 0.79 & 0.43 & 0.56 & 0.69 & 0.33 & 0.73 & 0.71 & 0.60 \\
PhyMind-base v2     & 0.72 & 0.72 & \textbf{0.90} & 0.81 & 0.80 & 0.83 & 0.42 & 0.86 & \textbf{0.84} & 0.77 \\
PhyMind-lite$^\ddagger$  & 0.93 & 0.92 & 0.95 & 0.89 & 0.87 & 0.88 & 0.50 & 0.92 & 0.89 & 0.86 \\
PhyMind-large$^\ddagger$ & 0.95 & 0.93 & 0.96 & 0.91 & 0.89 & 0.90 & 0.52 & 0.95 & 0.91 & 0.88 \\
MANAS-2 (released)  & \textbf{0.82} & \textbf{0.83} & 0.90 & \textbf{0.85} & \textbf{0.84} & \textbf{0.86} & 0.37 & \textbf{0.89} & 0.82 & \textbf{0.80} \\
\bottomrule
\end{tabular}
\end{table}

\subsection{How much does the physics path explain?}
We measured $R^2$ on held-out masked tokens of the full reconstruction $\hat{x}$, of the lead-field path $Gs$ alone, and of the residual $r$ alone. The phase-1 base models (v1, v2-H) have negative held-out $R^2$ ($-0.05$ to $-0.23$), i.e.\ their masked reconstructions are worse than predicting zero, even though they produce useful encoders. For lite and large the full reconstruction reaches 0.25/0.48 and 0.35/0.55 on EEGMAT/MI, of which $Gs$ alone gives 0.11/0.40 and 0.12/0.39; on two-channel Sleep-ST all models are at or below zero. The physical path thus carries a substantial share of the explained variance where reconstruction succeeds, but waveform reconstruction itself is weak, and the representational benefit appears to come mainly from the spectral likelihood rather than from accurate waveform prediction.

\subsection{Learned oscillators}
\begin{figure}[t]
  \centering
  \includegraphics[width=\textwidth]{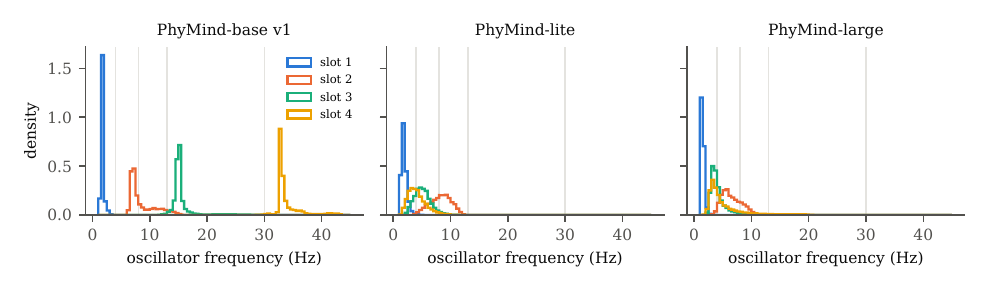}
  \caption{Distribution of the four oscillator-slot frequencies inferred on held-out data.}
  \label{fig:oscillators}
\end{figure}

In v1 the oscillator slots separate into roughly delta, theta, beta and gamma frequencies without supervision (Fig.~\ref{fig:oscillators}). With the stronger Whittle weight and the band head (lite, large) they collapse to 1.7--6\,Hz, so the ``four oscillators per region'' of the final models are not band-specific; the encoder and band head carry the higher-frequency information. A per-slot frequency prior would be needed to keep the slots separated.

\section{Limitations}
\begin{itemize}
  \item \textbf{Scale and seeds.} One pre-training seed per model, a corpus of 0.6--3k\,h, and test splits of 7 (EEGMAT) and 18 (ADFTD) subjects. Differences of a few points on these tasks are within noise; we do not have per-window predictions for a paired subject-level bootstrap.
  \item \textbf{Selection.} PhyMind variants were iterated on the evaluation tasks (Sec.~\ref{sec:setup}). The controlled baseline comparison predates this, but the lite/large results may be optimistically biased.
  \item \textbf{Scaling.} Base, lite and large differ in corpus, steps and objective version, so they do not form a clean scaling study.
  \item \textbf{Physics.} The lead field is a template head, the inverse problem is under-determined ($\le$64 channels, 68 regions), regional spectra are assumed independent, and the inferred sources do not localise motor imagery (Sec.~\ref{sec:localise}).
  \item \textbf{Comparison with MANAS-2.} We evaluate the released weights through our verified re-implementation with our own preprocessing (0.5--45\,Hz, bipolar channels placed at the anode), which may differ from its training pipeline. Our numbers for MANAS-2 on mental arithmetic and sleep staging are close to those on its model card.
\end{itemize}

\section{Conclusion}
Replacing the free-form decoder of an EEG masked autoencoder with a lead-field forward model driven by oscillator and aperiodic source dynamics substantially increases how much spectral structure is linearly decodable from encoder tokens, compared with plain masked reconstruction and with the MANAS-2 objective under identical conditions. This does not translate into large downstream gains: in controlled comparisons the physics objective is on par or slightly better, its region-level features help frozen motor-imagery decoding, and small PhyMind models are competitive with the released MANAS-2 when frozen but behind it after fine-tuning. The inferred sources are useful features but not valid localisations. Larger corpora, multiple seeds, and a mechanism that keeps sources anatomically faithful are the natural next steps.

\end{document}